\documentclass[10pt,twocolumn,letterpaper]{article}

\usepackage{iccv}
\usepackage{times}
\usepackage{epsfig}
\usepackage{graphicx}
\usepackage{amsmath}
\usepackage{amssymb}
\usepackage{amsfonts}
\usepackage{multicol}
\usepackage{multirow}
\usepackage{caption}
\usepackage{authblk}

\usepackage[pagebackref=true,breaklinks=true,letterpaper=true,colorlinks,bookmarks=false]{hyperref}
 \iccvfinalcopy 
\def\iccvPaperID{9523} 

\ificcvfinal\fi

\begin{document}
\title{DAVOS: Semi-Supervised Video Object Segmentation via Adversarial Domain Adaptation}
\setlength{\affilsep}{0pt}
\author{Jinshuo Zhang}
\author{Zhicheng Wang\thanks{Corresponding author.}}
\author{Songyan Zhang}
\author{Gang Wei}
\affil{CAD Research Center, Tongji University}
\affil{\textit{\tt\small\{zhangjinshuo, zhichengwang, spyder, weigang\}@tongji.edu.cn}}
\maketitle
\vspace{-10pt}
\ificcvfinal\thispagestyle{empty}\fi


\begin{abstract}
   Domain shift has always been one of the primary issues in video object segmentation (VOS), for which models suffer from degeneration when tested on unfamiliar datasets. Recently, many online methods have emerged to narrow the performance gap between training data (source domain) and test data (target domain) by fine-tuning on annotations of test data which are usually in shortage. In this paper, we propose a novel method to tackle domain shift by first introducing adversarial domain adaptation to the VOS task, with supervised training on the source domain and unsupervised training on the target domain. By fusing appearance and motion features with a convolution layer, and by adding supervision onto the motion branch, our model achieves state-of-the-art performance on DAVIS2016 with 82.6\% mean IoU score after supervised training. Meanwhile, our adversarial domain adaptation strategy significantly raises the performance of the trained model when applied on FBMS59 and Youtube-Object, without exploiting extra annotations.
\end{abstract}

\section{Introduction}
\label{secintro}
\begin{figure}
\begin{center}
   \includegraphics[width=0.8\linewidth]{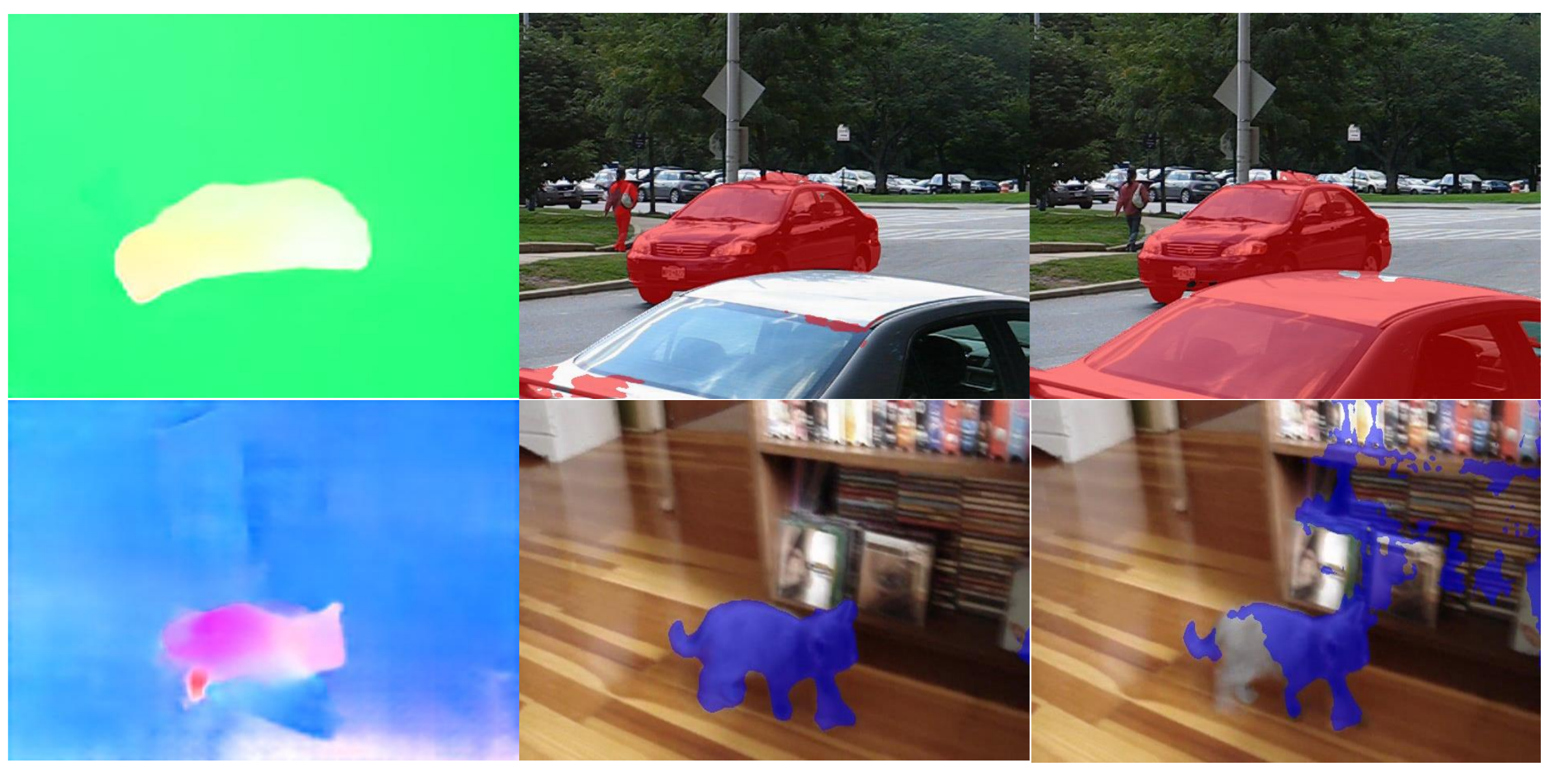}
\end{center}
\vspace{-3mm}
   \caption{Predictions between the model trained with and without our UDA method on FBMS59 \cite{fbms} and Youtube-Object \cite{ytb}. The optical flow (left) indicate the motion of objects. The results after UDA (middle) show that the model correctly recognizes the actual moving object (the car in the back) and produces less noise comparing to results before UDA (right).}
\label{intro}
\end{figure}

Detection of moving objects through image sequences can provide significant help for tasks like scene understanding and semantic analysis. But unlike semantic segmentation, video object segmentation (VOS) demands that the segmented object should be also in motion. Since it is not easy to tell whether an object is moving or not only depending on a single image, producing annotations for VOS tasks by hand can be cumbersome, which leads to the shortage of labeled scenes for supervised training. For that reason, the well-trained models on one dataset do not perform necessarily well when applied to another. As is stated in \cite{msk,uostr,mpnet} The issue of degeneration in VOS models has existed for a while, especially for off-line methods where the prior knowledge of target datasets is usually not leveraged during the training process. A major cause of such phenomenon is recognized as domain shift.

Efforts have already been made to reduce the harmful effects of domain shift and there is one major method called the domain adaptation (DA) strategy. The main idea behind DA is to align the data distribution of the source and the target domains in the latent space. Such a strategy has already been applied to many tasks like image classification \cite{da,adda,domainsym,udasimilarity} and semantic segmentation \cite{addressda4ss,advent} and been proved to be effective. However, not so much attention has been paid toward video object segmentation, as far as we know. For VOS tasks, there have been online methods called one-shot VOS \cite{osvos,eosvos,msk} that fine-tune on the annotation of the first frame in test sequences to gain performance, yet the annotations of test data are not always available to conduct such methods. Moreover, some unsupervised methods \cite{uvosmp, uovos, bnn, umod} are proposed to enable training on unlabeled datasets but there is still some work to be done to improve the overall accuracy and the performance under complicated scenes.

In this paper, we propose a method of a two-stream architecture based on deep convolutional networks, with unsupervised domain adaptation (UDA) involved to maintain the performance of the trained model on unlabeled datasets. The details of our methods are illustrated in \S \ref{secapp}. Evaluation have been conducted on DAVIS2016 \cite{davis}, FBMS59 \cite{fbms} and Youtube-Object \cite{ytb}, with DAVIS2016 treated as the source domain and the latter two as the target domain. We perform supervised training on DAVIS2016 and then using UDA strategy to minimize the discrepancy between the source domain and the target domain. Results show that we achieve state-of-the-art performance with mean intersection-over-union (IoU) score of $82.6\%$ on DAVIS2016. And by applying our UDA method to the model trained on the source domain, the IoU score of FBMS59 rises from $71.1\%$ to $75.6\%$ and the score of Youtube-Object rises from $59.8\%$ to $66.0\%$. Figure \ref{intro} provides examples that the model after domain adaptation provides better predictions on FBMS59 and Youtube-Object with less noise and fewer false positive results (the static car in the front). In summary, our main contributions include:
\begin{itemize}\item[$\bullet$] A video object segmentation model that effectively combines appearance and motion cues with a convolution layer. And through additional supervision on the motion branch, we make sure that the model distinguishes between moving and static objects with fewer false positive predictions.
\item[$\bullet$] Applying adversarial domain adaptation techniques to the VOS task with a domain confusion loss for unsupervised training on the target domain. The proposed UDA method remarkably reduces the side effects of domain shift for the model trained on the source domain, and provides segmentation masks with less noise and more proper boundaries without online fine-tuning.
\end{itemize}

\begin{figure*}
\centering\setlength{\belowcaptionskip}{-12pt}
\includegraphics[width=.70\textwidth]{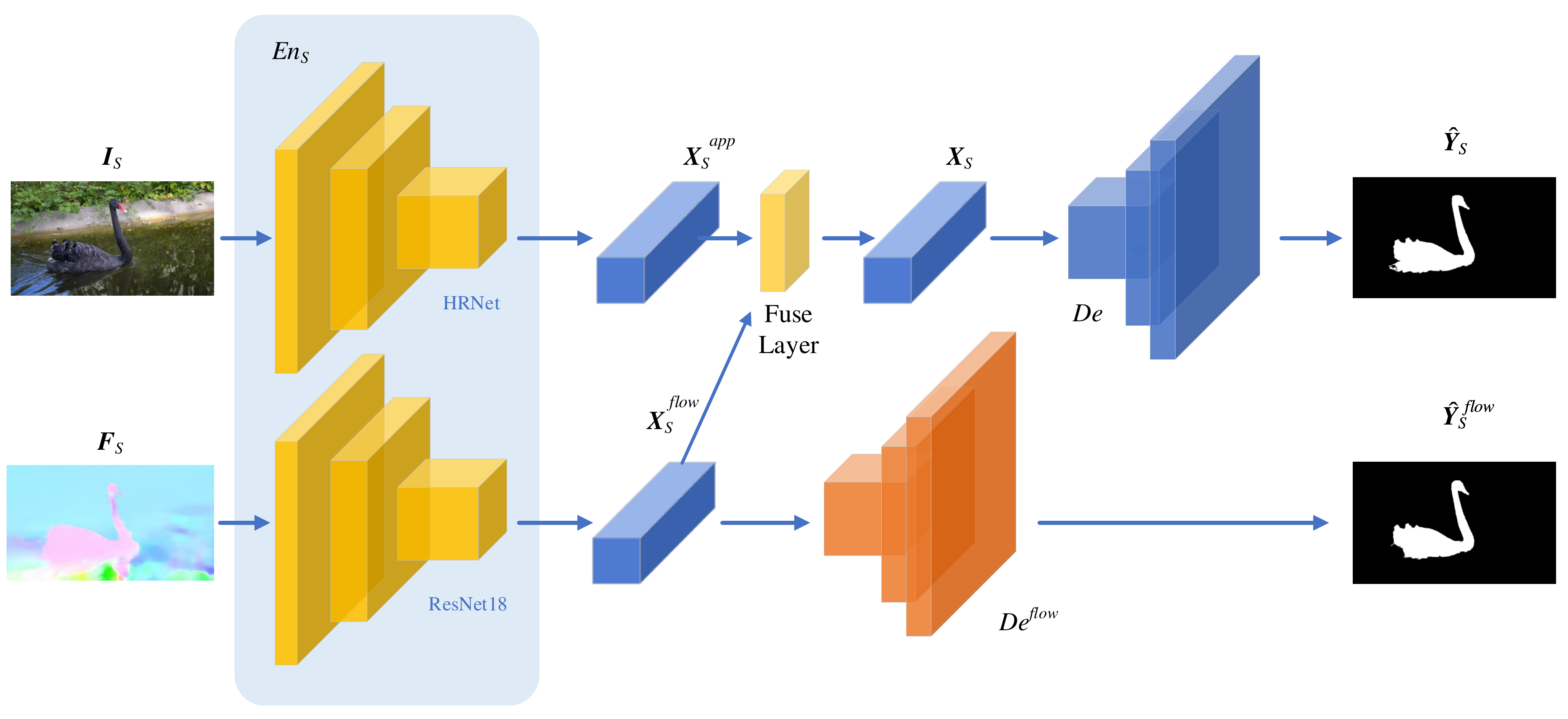}
   \caption{The supervised pipeline for the source domain. We utilize \cite{pwc} to produce optical flow $\boldsymbol{F}_S)$ as the motion cue and an RGB image as the appearance cue which are fed separately into the encoder $En_S$ . The appearance branch uses \cite{hrnet} as the backbone. And the flow branch exploits \cite{resnet} for feature extraction. $\boldsymbol{X}_S^{app}$ and $\boldsymbol{X}_S^{flow}$ meet at the fusion layer, which is implemented as a convolution module in the final version after comparing different operations (in Table \ref{davisablation}). The fused stream is then upsampled by $De$ multiple times to form a full-resolution segmentation mask. At the training stage, we conduct an additional mask prediction for the motion stream with a separated decoder $De^{flow}$, and the output is also monitored by the ground-truth.}
\label{supervise}
\end{figure*}

\section{Related work}\label{secrw}
\subsection{Video object segmentation}\label{secrw-vos}

\textbf{Online methods}
To preserve the robustness of the trained model, there are some strategies to apply the prior knowledge of the test data before testing, which is known as online training. A frequently leveraged one of such strategies is known as one-shot VOS \cite{osvos,onavos,eosvos,vosgm}. \cite{osvos} first introduces the one-shot concept on deep-learning-based VOS tasks. Its main attempt is to utilize the label of the first frame of test sequences, and perform supervised fine-tuning before making predictions for the whole sequence. \cite{msk} also utilizes online training as fine-tuning on test data, except that it also leverages the segmentation labels of the last frame when conducting training and testing on the current frame, achieving VOS on static images. \\

\textbf{Off-line methods}
Even though online training manages to mitigating the degeneration of the trained model, it relies on labels of test sequences, which are usually not available. For that reason, existing offline methods \cite{fseg,arp,ags,agunet,bnn,mpnet,lvosvm} attempt to make better use of available resources. As illustrated in \S \ref{secintro}, the VOS task aims at detecting moving foreground in images so that the information about motion is often needed. As is exploited in many methods, the optical flow becomes a popular alternative as a motion cue. \cite{mpnet} succeeds in performing VOS merely through optical flow computed with consecutive frames from videos. \cite{mod} provides better results through two-stream (motion stream and appearance stream) convolutional networks to extract features from optical flow and RGB images separately, and train the model to perform motion segmentation and object detection at the same time for more generalized feature representations. \cite{fseg} extends the two-stream idea with a late fusion strategy with which the motion stream and appearance stream meet right before the pixel-wise classification, \ie, the segmentation layer. The motive behind this is to prevent the tendency of the appearance stream being dominant, which is a noticed issue of the original two-stream architecture. In addition to supervised methods, there are also many unsupervised methods \cite{bnn, uvosva, uvosmp} that work well without annotations provided. Moreover, new techniques have been introduced to VOS recently, such as visual attention \cite{cosnet, uvosva} and graph neural networks \cite{agnn}.

To leverage the nature of videos, some methods pay attention to the spatio-temporal context. \cite{lvosvm} applies GRU layers to propagating initial hidden states over time. Likewise, there have been methods utilize various form of representation when conducting propagation. \cite{bnn} uses graph cut on bilateral filtered vectors, but unlike previous work, the graph is built from a small subset of pixels instead of the whole image, preserving efficiency and accuracy. \cite{uvosmp} choose to use superpixels as nodes when building graphs with spatio-temporal connections among nodes. \cite{arp} also uses superpixels but implements the method with the assumption that the object candidates reoccurring through the image sequence are likely to be objects of interest, and perform matching of candidates when refining the segmentation results. 

\subsection{Unsupervised domain adaptation}\label{secrw-uda}
To define the divergence between the distribution of source domain and the target domain, the maximum mean discrepancy (MMD) \cite{mmd} is proposed to measure the distance between these two distributions by difference of mean values. Then there is the adversarial domain adaptation. \cite{da} applies the adversarial training onto DA tasks by introducing a gradient reversal layer so that the model can be trained with a standard forward and backward propagation manner. Based on \cite{da}, \cite{adda} further extends the idea of adversarial UDA by comparing the major factors that affect the performance of different UDA methods and uses domain confusion loss during training. For computer vision tasks, the UDA is first tested on image classification tasks, such as \cite{da,adda,udasimilarity,progressiveda,adadm}. Other tasks like semantic segmentation also notice the impact of DA \cite{clan,advent,addressda4ss,bidirectional,contextualda}. \cite{clan} notices the shortage of global domain adaptation ignores the semantic consistency among categories in tasks like semantic segmentation. It performs category-level alignment by overlooking well-aligned classes and focusing on more differently-distributed classes between source and target domains in semantic segmentation. \cite{advent} introduces a pixel-wise entropy minimization plot to penalize the low-confidence predictions on the target domain. In \cite{contextualda}, observation is made that while in supervised semantic segmentation the contextual relations were exploited, it did not appear in UDA methods. And it proposes to align features by local-level consistencies, modeling the relation by clustering patches sampled from label maps by gradient histogram. Another strategy in UDA is to set up a prototype. \cite{udasimilarity} replaces typical modules like fully-connected layers and softmax with a similarity calculation when doing image classification, assuming there exists a prototype for each category. Such pivot aims to raise the robustness against noise correlated with shared distribution.

\begin{figure*}
\centering\setlength{\belowcaptionskip}{-12pt}
\includegraphics[width=.65\textwidth]{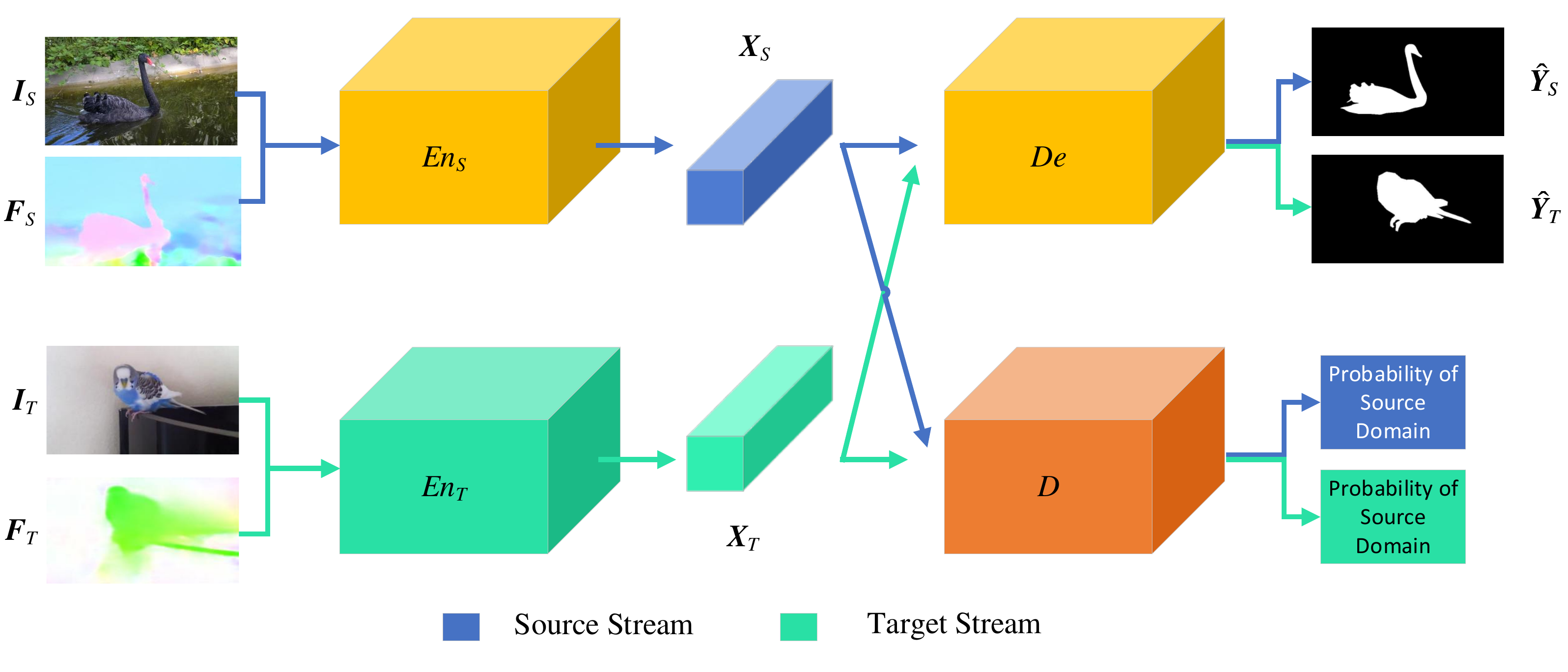}
\vspace{-4mm}
   \caption{The pipeline of our weights-separated UDA method. The blue stream represents the forward propagation on the source domain with weights of $En_S$ and $De$ fixed throughout the training. The green stream shows the separated encoder $En_T$ generating features exclusively for the target domain. During training, features from both domains, \ie, $\boldsymbol{X}_S$ and $\boldsymbol{X}_T$ are classified by discriminator $D$. The output probability of $D$ indicates which domain the input feature belongs to. When optimizing $D$, the label of source and target domain are set as $1$ and $0$, but when optimizing $En_T$ the labels switch to achieve domain confusion.}
\label{unsupervise}
\end{figure*}


\section{Approach}\label{secapp}
We propose a framework to perform supervised learning for video object segmentation on labeled datasets (the source domain) and maintain the performance of the model when applied to test data (the target domain) by unsupervised domain adaptation. The main idea of UDA is to bridge the gap between the distributions of different domains, \eg, different scenes or categories of objects by training a discriminator with domain confusion loss.
\subsection{Overview}\label{secapp-overview}
We denote the images from the source domain as $\boldsymbol{I}_S\in\mathbb{R}^{H\times W\times 3}$, and images from the target domain as $\boldsymbol{I}_T\in\mathbb{R}^{H\times W\times 3}$. Optical flow $\boldsymbol{F}\in\mathbb{R}^{H\times W\times 2}$ is generated from consecutive images of sequences as $\boldsymbol{F}=O(\boldsymbol{I}^t,\boldsymbol{I}^{t-1})$, where $O$ can be replaced with many methods. In our work, we utilize a pretrained PWC \cite{pwc} network for flow extraction. Flow maps are viewed as the motion cue for both the source and target domains. Also, we denote the predicted segmentation masks as $\boldsymbol{\hat{Y}}\in\mathbb{R}^{H\times W\times 1}$ and ground-truth masks as $\boldsymbol{Y}\in\mathbb{R}^{H\times W\times 1}$ The segmentation network consists of two sub-modules denoted as encoder $En(\cdot)$ and decoder $De(\cdot)$. The idea behind such a setting is that the alignment for distributions of domains is processed in the feature space instead of the final segmentation outputs. In that case, the aligned features should be sharing the same output layers, \ie, the decoder, to get the segmentation results as (\ref{eq1}) and (\ref{eq7}). During the supervised learning phase, the encoder and decoder are trained in an end-to-end fashion. For UDA learning, we use two setups in which one conducts supervised training on source domain and UDA on target domain at the same time with the same encoder. The other utilizes another encoder for target domain and perform UDA after supervised training.
\subsection{Supervised dual-branch VOS}\label{secapp-su}
As is presented in Figure \ref{supervise}, the encoder network utilized the dual-branch architecture of \cite{mod}, where the appearance features and motion features are generated separately by feeding a single image and its flow map as (\ref{eq3}). We replace the feature extractor of the appearance stream and motion stream with HRNet\cite{hrnet} and ResNet\cite{resnet}. HRNet runs parallel feature extraction on different scales which boosts the performance of many tasks. For the fusion of those streams, we have tried multiple operations, such as pixel-wise product, pixel-wise addition, as well as a convolution layer. The comparison among different fusion methods can be found in \S \ref{secex-ablation}. The forward-propagation steps are listed as follows:
\begin{equation}
\label{eq3}
[\boldsymbol{X}_S^{app}, \boldsymbol{X}_S^{flow}]=En_S([\boldsymbol{I}_S,\boldsymbol{F}_S])
\end{equation}
\begin{equation}
\label{eq0}
\boldsymbol{X}_S=\phi([\boldsymbol{X}_S^{app}, \boldsymbol{X}_S^{flow}])
\end{equation}
\begin{equation}
\label{eq1}
\boldsymbol{\hat{Y}}_S=De(\boldsymbol{X}_S)
\end{equation}

\begin{table*}[t]
\setlength{\belowcaptionskip}{-12pt}
\centering\resizebox{\textwidth}{10mm}{
    \begin{tabular}{|c|c||c|c|c||c|c|c|c||c|c|c|c|c|c|c|c|}
    \hline
    \multicolumn{2}{|c||}{}&\multicolumn{3}{c||}{Unsupervised} & \multicolumn{4}{c||}{Supervised-Online} & \multicolumn{8}{c|}{Supervised-Offline}\\ \cline{3-17}
    \multicolumn{2}{|c||}{Metrics} &UVOSBN\cite{bnn}  &UVOSVA\cite{uvosva}& UOVOS\cite{uovos}&OnAVOS\cite{onavos}&OSVOS\cite{osvos}&eOSVOS\cite{eosvos}&MSK\cite{msk}&VOSGM\cite{vosgm}&FSEG\cite{fseg}&ARP\cite{arp}  & MP-Net\cite{mpnet} &AGNN\cite{agnn}&LVO\cite{lvosvm}&COSNet\cite{cosnet}& \textbf{Ours}\\ \hline
    \multirow{3}{*}{$\mathcal{J}$}  &Mean $\mathcal{M}\uparrow$ &\textbf{80.4}    &79.7         &77.8         &86.1         &79.8         & \textbf{86.6}&79.7  &82.5           & 70.7 & 76.3 & 69.7 &80.7         & 75.9       &80.5         &\textbf{82.6}\\ \cline{2-17}
                                    & Recall $\mathcal{O}\uparrow$       &93.2    &91.1         &\textbf{93.6}&\textbf{96.1}& 93.6        & --           &93.1  &\textbf{94.3}& 83.5   & 91.1 & 82.9 &94.0         &89.1        &94.0         &93.0         \\ \cline{2-17} 
                                    & Decay $\mathcal{D}\downarrow$     &4.8      &\textbf{0.0} &2.1          &5.2          & 14.9        &\textbf{4.5}  &8.9   &4.2            & 1.5  & 7.0  &5.6   &0.03         &\textbf{0.0}&\textbf{0.0} &7.4          \\ \hline
    \multirow{3}{*}{$\mathcal{F}$}  & Mean$\mathcal{M}\uparrow$     &\textbf{78.5}&77.4         &72.0         &84.9         & 80.6        &\textbf{87.0} &75.4  &81.2           & 65.3 & 70.6 & 66.3 &79.1         & 72.1       &79.4         &\textbf{83.9}\\ \cline{2-17} 
                                    & Recall $\mathcal{O}\uparrow$  &\textbf{88.6}&85.8         &87.7         &89.7         &\textbf{92.6}& --           &87.1  &90.3           & 73.8 & 83.5 & 78.3 &\textbf{90.5}& 83.4       &90.4         &90.2         \\ \cline{2-17} 
                                    & Decay $\mathcal{D}\downarrow$         &4.4  &\textbf{0.0} &3.8          &\textbf{5.8} & 15.0        & --          &9.0    &5.6            & 1.8  & 7.9  &6.7   &0.03         & 1.3        &\textbf{0.0} &4.3          \\ \hline
    \end{tabular}
}

\caption{The results shown above are all produced on the DAVIS2016 validation split. We group those results into three groups of unsupervised, supervised-online (one-shot), and supervised-offline. Here we present our results in the supervised-offline group, since we evaluate our model on the test set without further online fine-tuning. It shows that our proposed method outperforms other supervised offline methods in as well as unsupervised methods in mean $\mathcal{J}$ score and mean $\mathcal{F}$ score, and produces competitive results when compared to online methods.}
\label{davisbenchmark}
\end{table*}

 $\phi$ is the fusion layer. We found that simply fuses the two streams prevent the model from better performance since the model tends to amplify the influence of the appearance cue and the motion cue is poorly leveraged. Consequences are that the model classifies all specific objects, \eg camels and people, into regions of interest regardless they are moving or not (shown in Figure \ref{fig:daviscompare}). To tackle this issue, we set up constraints on the flow branch. During training, the main segmentation is performed by the decoder which takes the fused feature representations and produces masks $\boldsymbol{\hat{Y}}_S$. The binary cross-entropy loss (\ref{eq4}) is applied to $\boldsymbol{\hat{Y}}_S$ to conduct supervised learning; Meanwhile, another decoder dedicated for motion feature also produces masks $\boldsymbol{\hat{Y}}_S^{flow}$ as (\ref{eq2}) and is supervised by another binary cross-entropy loss as well and the total loss is produced as (\ref{eq5}). Unlike \cite{fseg} that performs late fusion for three streams, in our model, only the fused branch and flow branch are to produce masks and be constrained at the training phase. The decoder itself consists of repeated convolution and upsampling layers and no skip-connections are exploited between encoder and decoder. The final output segmentation masks are of the same resolution as the input images. 
 \begin{equation}
\label{eq2}
\boldsymbol{\hat{Y}}_S^{flow}=De^{flow}(\boldsymbol{X}_S^{flow})
\end{equation}
\begin{equation}
\label{eq4}
L_{msk}(\boldsymbol{Y},\boldsymbol{\hat{Y}})=\mathbb{E}[\boldsymbol{Y}(\boldsymbol{\hat{Y}})+(1-\boldsymbol{Y})log(1-\boldsymbol{\hat{Y}})] 
\end{equation}
\begin{equation}
\label{eq5}
L_S=\alpha_1L_{msk}(\boldsymbol{Y}_S,\boldsymbol{\hat{Y}}_S)+\alpha_2L_{msk}(\boldsymbol{Y}_S,\boldsymbol{\hat{Y}}_S^{flow})
\end{equation}

$\alpha_1$ and $\alpha_2$ are leveraged to balance the optimization process to make better use of the motion cue. $\mathbb{E}$ stands for expectation.
The trained model is able to provide full-resolution masks for images of random sizes. And in the test phase, we take advantage of fully-connected CRFs \cite{crf} to further augment the predictions. 

\subsection{Adversarial domain adaptation}\label{secapp-ada}
Now we obtain the well-trained model which is able to perform VOS and provide qualitative results for source domain. However, it does not mean that it can also generalize well on other datasets, which is caused by domain shift as we have previously illustrated. To address this issue, we exploit the UDA method through training the feature extractor to offer representations that blurs the boundaries between source and target domains, meanwhile no segmentation labels are needed, which is achieved with adversarial learning. It should be noted that among the adversarial UDA methods, choices are made about to what extent the weights of feature extractors for different domains should be shared. The source domain and the target domain can share the same feature extractor in some cases \cite{advent,da}, or with weights partly shared \cite{partlyshared}, or even completely separated \cite{adda}. To find the exact architecture that solving the domain shift at best for the current task, evaluation is conducted for two setups which we call the weights-shared and weights-separated UDA strategies.

Specifically, the weights-separated method uses separated encoders for different domains as is shown in Figure \ref{unsupervise}. $En_S$ and $De$ trained on the source domain are both frozen at this stage, and $En_S$ is only used to perform forward propagation for input data from the source domain and produce the features $\boldsymbol{X}_S$. Then we set up another encoder $En_T$ for data from the target domain and denote its output features as $\boldsymbol{X}_T$. The decoder $De$ serves both $\boldsymbol{X}_S$ and $\boldsymbol{X}_T$ in to produce segmentation masks. The forward propagation of data from the target domain is listed as follows:
\begin{equation}
\label{eq6}
\boldsymbol{X}_T=\phi(En_T([\boldsymbol{I}_T, \boldsymbol{F}_T]))
\end{equation}
\begin{equation}
\label{eq7}
\boldsymbol{\hat{Y}}_T=De(\boldsymbol{X}_T)
\end{equation}

As for the domain confusion loss (\ref{eq8}), it targets at minimizing the discrepancy between $\boldsymbol{X}_S$ and $\boldsymbol{X}_T$ by optimizing $En_T$ to fool the discriminator $D$ from correctly distinguishing the domain to which any input feature belongs. At the same time, $D$ must upgrade itself with (\ref{eq9}) to make better decisions against the domain alignment conducted by $En_T$. 
\begin{equation}
\label{eq8}
L_{En_T}=-\mathbb{E}[log(D(\boldsymbol{X}_T))]
\end{equation}
\begin{equation}
\label{eq9}
L_{D}=-\mathbb{E}[log(D(\boldsymbol{X}_S))]-\mathbb{E}[log(1-D(\boldsymbol{X}_T))]
\end{equation}
\begin{equation}
\label{eq10}
L_{UDA}=\beta_1L_{En_T}+\beta_2L_{D}
\end{equation}
Those illustrated above are exactly the idea of adversarial learning. During the training phase, the procedure can be divided into two steps. We first freeze the $En_S$ and $En_T$ and train $D$ by generating features from source and target domains and feed them to the discriminator separately to get the classification values. The loss for $D$ is calculated as (\ref{eq9}). We denote labels of source domain or target domain as 1 and 0 here. After backpropagation for $D$, there is a second step where we train the $En_T$ with only input data from the target domain and no segmentation labels are required. Still, $D$ provides a classification value for the output $\boldsymbol{X}_T$. We calculate the loss for $En_T$ as (\ref{eq8}) with the labels for target domain reset as 1, which is why it is called domain confusion loss. The weights of $D$ freeze at this step and optimization is done only for $En_T$. We repeat these two steps until domain alignment is reached, the overall loss in this stage is formed as (\ref{eq10}).
\begin{equation}
\label{eq11}
L_{UDA-shared}={L_S}+\lambda_1L_{En_T}+\lambda_2L_D
\end{equation}
Things get easier for the weights-shared setup as it shares the same encoder $En_S$ and UDA is performed along with the supervised training but without using labels for data of the target domain. The discriminator $D$ is also exploited the same way as previously stated. During training, we first run forward and backward propagation for $\boldsymbol{I}_S$ and $\boldsymbol{F}_S$ for one iteration with $L_S$ and then replace the inputs with $\boldsymbol{I}_T$ and $\boldsymbol{F}_T$ and the forward propagation stops after $\boldsymbol{X}_T$ is produced. The loss for target domain is  also calculated as (\ref{eq8}). The loss at this stage is formed as (\ref{eq11}) After backward propagation for the target domain, we run training for the discriminator to improve its ability for domain classification. With such strategy, the model after training works well on both the source domain and the target domain.

\section{Experiments}\label{secex}
\begin{table*}
\centering\setlength{\belowcaptionskip}{-2pt}
\scriptsize
\begin{tabular}{|c|c|c|c|c|c|c|c|c|}
\hline
Method & FSEG\cite{fseg}& MSK\cite{msk} & BVS\cite{bvs} & CoSeg \cite{coseg}& FCN\cite{fcn}& \textbf{Ours}(DA-Shared) & \textbf{Ours}(DA Separated) & \textbf{Ours}(w/o DA)\\ \hline
$\mathcal{J}$ Mean  & 68.6& \textbf{71.7} & 59.7& 62.3  & 56.9& 66.0 &62.2 &59.8        \\ \hline
\end{tabular}
\caption{Evaluation on the Youtube-Object dataset are conducted on its test split. The $\mathcal{J}$ mean score of our shared UDA method is at the third place among the methods listed above, and our DA method with shared weights drives the score up from $59.8\%$ to $66.0\%$.}
\label{ytbbenchmark}
\end{table*}

\begin{table*}[]
\centering\scriptsize\setlength{\belowcaptionskip}{-12pt}
\begin{tabular}{|c|c|c|c|c|c|c|c|c|c|}
\hline
\multicolumn{10}{|c|}{$\mathcal{J}$ Mean Score per Scene on DAVIS Test Split}                \\ \hline
Scene               & OnAVOS\cite{onavos}& ARP\cite{arp}  &OSVOS\cite{osvos}& MSK\cite{msk} & LVO\cite{lvosvm} &UOVOS\cite{uovos} & FSEG\cite{fseg} & COSNet\cite{cosnet} & \textbf{Ours}    \\ \hline
blackswan           & \textbf{96.3} & 88.1 & 94.2 & 90.3&          74.1& 70.5 & 81.2 &  88.0  & 93.3             \\ \hline
bmx-trees           & 58.1          & 49.9 & 55.5 & 57.5&          49.9& 56.6 & 43.3 &  46.5  & \textbf{60.9}             \\ \hline
breakdance          & 70.9 & \textbf{76.2} & 70.8 & \textbf{76.2}& 37.1& 60.1 & 51.2 &  68.3  & 53.6             \\ \hline
camel               & 85.4 & 90.3 & 85.1 & 80.1&                   88.1& 76.0 & 83.6 &  89.4  & \textbf{94.0}             \\ \hline
car-roundabout      & \textbf{97.5} & 81.6 & 95.3 & 96.0&          88.6& 89.4 & 90.7 &  94.7  & 86.6             \\ \hline
car-shadow          & \textbf{96.9} & 73.6 & 93.7 & 93.5&          92.0& 95.0 & 89.6 &  93.5  & 95.9            \\ \hline
cows                & \textbf{95.5} & 90.8 & 94.6 & 88.2&          90.2& 85.2 & 86.9 &  91.4  & 95.4             \\ \hline
dance-twirl         & \textbf{84.4} & 79.8 & 67.0 & \textbf{84.4}& 81.0& 70.9 & 70.4 &  77.7  & 83.5             \\ \hline
dog                 & \textbf{95.6} & 71.8 & 90.7 & 90.9&          88.7& 86.1 & 88.9 &  93.7  & 94.7             \\ \hline
drift-chicane       & \textbf{89.2} & 79.7 & 83.5 & 86.2&          63.9& 38.9 & 59.6 &  70.5  & 73.2             \\ \hline
drift-straight      & \textbf{94.4} & 71.5 & 67.6 & 56.0&         84.9& 81.1 & 81.1 &  91.7  & 91.8             \\ \hline
goat                & \textbf{91.3} & 77.6 & 88.0 & 84.5&          82.3& 86.6 & 83.0 &  83.6  & 88.4             \\ \hline
horse-jump-high     & \textbf{90.1} & 83.8 & 78.0 & 81.7&          82.4& 79.7 & 65.2 &  81.9  & 81.8             \\ \hline
kite-surf           & 69.1 & 59.1 & 68.6 & 60.0&                   64.6& 62.9 & 39.2 &  67.5  & \textbf{76.9}             \\ \hline
libby               & \textbf{88.4} & 65.4 & 80.8 & 77.5&          69.0& 75.6 & 58.4 &  69.0  & 74.6             \\ \hline
motocross-jump      & 82.3 & 82.3 & 81.6 & 68.5&                   80.5& 70.8 & 77.5 &  82.5  & \textbf{85.9}             \\ \hline
paragliding-launch  & 64.3 & 60.1 & 62.5 & 62.0&                   62.2& 61.4 & 57.1 &  61.2  & \textbf{65.2}             \\ \hline
parkour             & \textbf{93.6} & 82.8 & 85.6 & 88.2&          84.9& 85.9 & 76.0 &  87.7  & 92.9             \\ \hline
scooter-black       & \textbf{91.1} & 74.6 & 71.1 & 82.5&          71.8& 72.4 & 68.8 &  83.8  & 74.5            \\ \hline
soapbox             & 88.5 & 84.6 & 81.2 & \textbf{89.9}&          81.3& 71.9 & 62.4 &  87.3  & 89.3             \\ \hline
\end{tabular}
\caption{We provide $\mathcal{J}$ mean score per scene on DAVIS test split. The table shows that we make the best predictions on second-most scenes, only fewer than \cite{onavos} which exploit online fine-tuning on test data. }
\label{davisperseq}
\end{table*}

\begin{figure*}
\begin{center}\setlength{\belowcaptionskip}{-12pt}
\includegraphics[width=.65\textwidth]{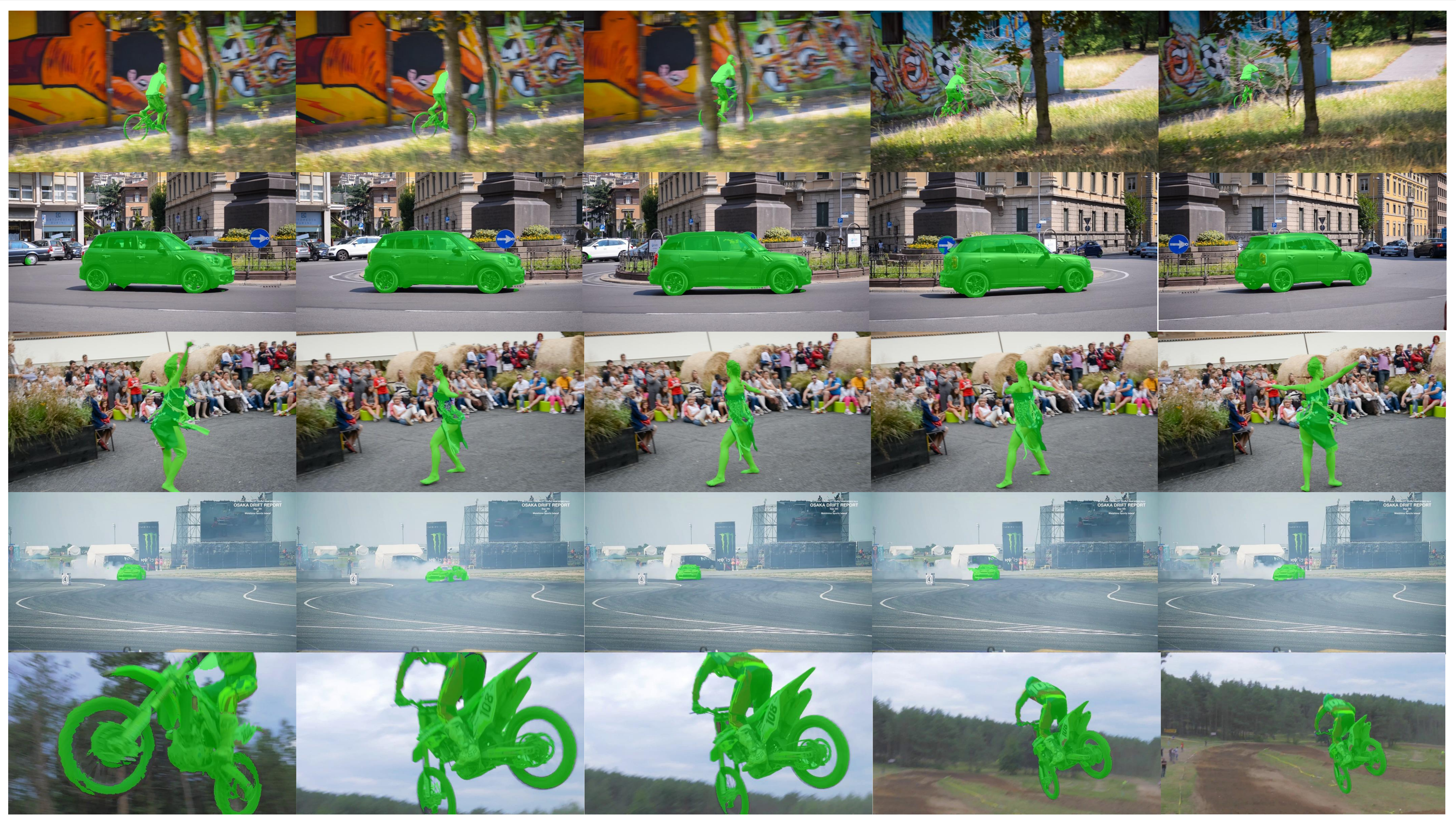}
\end{center}
\vspace{-5mm}
  \caption{Qualitative results of our supervised method on the DAVIS validation split. The chosen scenes represent tough occasions, such as occlusion, shadows, non-rigid motion, small objects and complicated shapes. drastic change in shape, which our method manages to handle.}
\label{davisquality}
\end{figure*}

\begin{table*}
\scriptsize\centering\setlength{\belowcaptionskip}{-12pt}
\begin{tabular}{|c|c|c|c|c|c|c|c|c|}
\hline
Method            & ARP\cite{arp} & UMOD\cite{umod} & UOVOS\cite{uovos} &UVOSMP\cite{uvosmp}& UVOSBN\cite{bnn} & \textbf{Ours}(DA Shared) & \textbf{Ours}(DA Separated) & \textbf{Ours}(w/o DA) \\ \hline
$\mathcal{J}$ Mean & 59.8          & 63.6            &  63.9                &60.8               &73.9            &   \textbf{75.6}      & 72.7                 &71.1      \\ \hline
\end{tabular}
\caption{Evaluation of our UDA methods on the test split of FBMS59. Our methods provide competitive results with the state-of-the-art methods and compared to the result produced with model trained on source domain, our DA method raises the score by a lot, especially with the pipeline of shared weights which earns $75.3\%$ for the mean IoU score.}
\label{fbmsbenckmark}
\end{table*}
\subsection{Dataset}\label{secex-dataset}
\textbf{DAVIS2016} \cite{davis} contains 50 video sequences of various scenes. It also provides fine-grained object mask annotations for each frame. There are 3455 frames in total with 2079 for training and 1376 for test. There are also two options for the image resolution, and we use the 480p version in our experiments.

\textbf{FBMS59} \cite{fbms} provides 59 video sequences of different scenes with 30 sequences for test, but only a small subset of frames in each sequence are annotated. We run UDA on its training split and conduct evaluation on the test split.

\textbf{Youtube-Object} \cite{ytb} includes videos with 10 object categories but 126 sequences in total which are not fully-annotated, either. Evaluation is performed on the ground truth provided by \cite{ytbann}.

\subsection{Implementation details}\label{secex-imp}
\textbf{Network architecture} The main model is implemented with an an encoder-decoder architecture. For the encoder, HRNet \cite{hrnet} and ResNet18 \cite{resnet} are adopted for the backbones of appearance branch and the flow branch separately. We test three operations (convolution, pixel-wise product and addition) for fusion of those two branches. For the decoder, we form a module with stacked convolution and bilinear interpolation layers to upsample the fused features. Final predictions share the same resolution as input images. Our methodology of UDA is conducted in two ways. The one with weights separated demands the training divide into a supervised phase and UDA phase. The supervised training is conducted with the model illustrated above but the unsupervised pipeline requires a clone of $En_S$ and a discriminator for adversarial learning. The discriminator is set as a full-convolution network which produces a scalar value for each input sample indicating its domain. The other UDA scenario conduct UDA and supervised learning at the same time so that $En_T$ is not required in this setup but the discriminator is. 
\begin{table*}[]
\centering\setlength{\belowcaptionskip}{-12pt}
\scriptsize
\begin{tabular}{|c|c|c|c|c|c|c|c|c|}
\hline
Module            & Flow-Branch&Product-Fusion&Addition-Fusion&Conv-Fusion & Flow-Supervision& $\mathcal{J}$ Mean & $\mathcal{J}$ Mean (w/ CRFs)\\ \hline
Baseline          &            &              &               &            &                 & 77.6               &     78.1                 \\ \hline
Baseline+FB+PF    &  $\surd$   &   $\surd$    &               &            &                 & 79.0               &     79.6                 \\ \hline
Baseline+FB+AF    &  $\surd$   &              &    $\surd$    &            &                 & 79.1               &     79.9                 \\ \hline
Baseline+FB+CF    &   $\surd$  &              &               &  $\surd$   &                 & 80.2               &     80.5                 \\ \hline
Baseline+FB+CF+FS &   $\surd$  &   $\surd$    &               &            & $\surd$         & 79.4               &     80.3                 \\ \hline
Baseline+FB+AF+FS &   $\surd$  &              &    $\surd$    &            & $\surd$         & 80.4               &     80.9                 \\ \hline
Baseline+FB+AF+FS &   $\surd$  &              &               &  $\surd$   & $\surd$         & \textbf{81.8}      &     \textbf{82.6}        \\ \hline

\end{tabular}
\caption{The table provides comparisons among different options for supervised training on DAVIS2016. It shows that using the flow branch as motion cue provides significant help for the VOS task. Also, the convolution-based fusion strategy performs better than the other two operations and adding supervision onto the flow branch further improves the results. With fully-connected CRFs as post-process, we finally achieve mean IoU of $82.6\%$ on the test split.}
\label{davisablation}
\end{table*}

\textbf{Training}
 The supervised training on the source domain is performed in an end-to-end fashion by minimizing $L_S$, with $\alpha_1=0.5$ and $\alpha_2=0.5$. We use SGD as optimizer with momentum of $0.9$ and weight decay of $0.0001$. The initial learning rate is $0.004$ and decreases with exponential strategy after each epoch. Pretrained weights provided by \cite{hrnet} and \cite{resnet} are loaded before training. We exploit the DAVIS2016 train split as the source domain for training. Square patches of $384\times384$ are extracted with random crop on original images. Optical-flow is produced by PWCNet \cite{pwc} from those patches. Since optical flow only has two channels, we add a third channel with matrix of ones to adapt to ResNet. We perform horizontal flipping, color jittering as data augmentation as well. The training is done after 100 epochs with Nvidia GTX 2080 Ti. The weights-shared UDA method is also conducted at this stage and the optimizer for discriminator share the same settings as the one for supervised training except that the momentum is set to $0$. $\lambda_1$ is updated with $\frac{Epoch}{MAX Epoch}$ for each epoch and $\lambda_2$ is set to $0.5$ during training. The data of target domain includes training splits of FBMS59 and Youtube-Object. We perform UDA on these two datasets separately.

For weights-separated UDA, $En_T$ is initialized with the weights of $En_S$ after the supervised training stage. Throughout the UDA session, weights of $En_S$ and $De$ are fixed. We use SGD optimizers for both $D$ and $En_T$ with initial learning rate of $0.0001$. For each step, $D$ is first optimized for $N$ iterations on both features of source and target domains. Then $En_T$ is optimized for $M$ iteration on input data of target domain with $\beta_1=1.0$ and $\beta_2=0.5$ for loss production. Training is done after 20 epochs and we finally set $M=N=5$ after many attempts.

\subsection{Evaluation} We present the comparison between our method and other state-of-the-art methods of VOS tasks. The metrics include the mean Jaccard score (IoU) denoted as $\mathcal{J}$ and the $\mathcal{F}$ score indicating contour accuracy.

For evaluation on DAVIS2016, we use the model trained at the supervised stage without fine-tuning on the test split. Different from the training phase, we use images with full resolution and optical flow produced from images of $512\times512$. We also apply fully-connected CRFs \cite{crf} to the predictions as refinement. Table \ref{davisbenchmark} lists the accuracy of different methods with $\mathcal{J}$ and $\mathcal{F}$. Besides, we group methods into unsupervised, supervised-online, and supervised-offline for fair comparison. As is presented, our method outperforms other state-of-the-art supervised-offline methods as well as unsupervised methods with $82.6\%$ of mean $\mathcal{J}$ score and $83.9\%$ of mean $\mathcal{F}$ score, and provide comparable results in terms of online methods. Table \ref{davisperseq} provides comparison of scores on each scene. Our method achieves best scores of second most scenes among the methods listed. We also provide some qualitative results in Figure \ref{davisquality}. It shows that the model works well in specific tough scenes, such as occlusion, shadows, non-rigid motion, small objects, and complicated shapes. Further study in \S \ref{secex-ablation} reveals some of the main factors that raise the performance of our model.  

\begin{figure}
\centering\setlength{\belowcaptionskip}{-12pt}
\includegraphics[width=.9\linewidth]{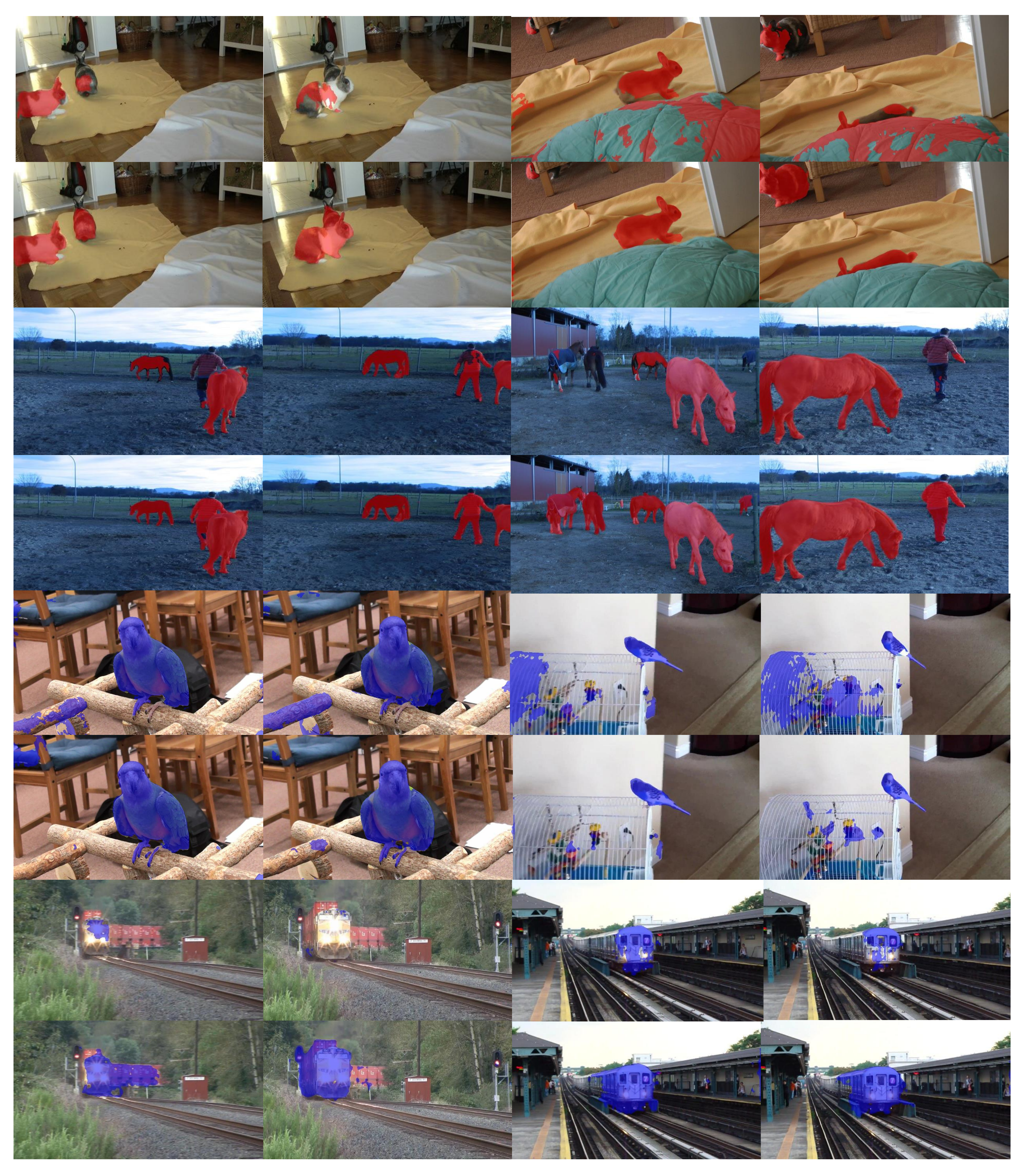}
   \caption{The figure shows comparison between predictions on target domain,\ie, FBMS59 and Youtube-Object. We use red masks for FBMS59 and blue masks for Youtube-Object. For each scene, the upper row shows results before UDA and the lower row shows the results after UDA. It clearly shows that, the output masks of model after UDA tend to have less noise without omitting moving objects (the person besides horses and the train). }
\label{fbmscomparison}
\end{figure}

For FBMS59 and Youtube-Object, we conduct evaluation after domain adaptation on each dataset. Results are provided in Table \ref{fbmsbenckmark} and Table \ref{ytbbenchmark}. The tables shows that the model after weights-shared UDA provide better results on FBMS59 \cite{fbms} than the other state-of-the-arts with $75.6\%$ mean $\mathcal{J}$ score. On Youtube-Object \cite{ytb}, the $\mathcal{J}$ score rises from $59.8\%$ to $66.0\%$ after UDA which increases by $10.4\%$. It shows that our proposed UDA methods manages to maintain the performance of the trained model when applied to unfamiliar datasets, \ie, the target domain. Figure \ref{fbmscomparison} shows the improvements our UDA method has made for some scenes of FBMS59 and Youtube-Objects. It can be seen that predictions per scene after UDA contain less noise and provide better boundaries for the moving objects. We further discuss the behaviors of different setups in our UDA method at \S \ref{secex-ablation}. 

\begin{figure}
\centering\setlength{\belowcaptionskip}{-12pt}
\includegraphics[width=.9\linewidth]{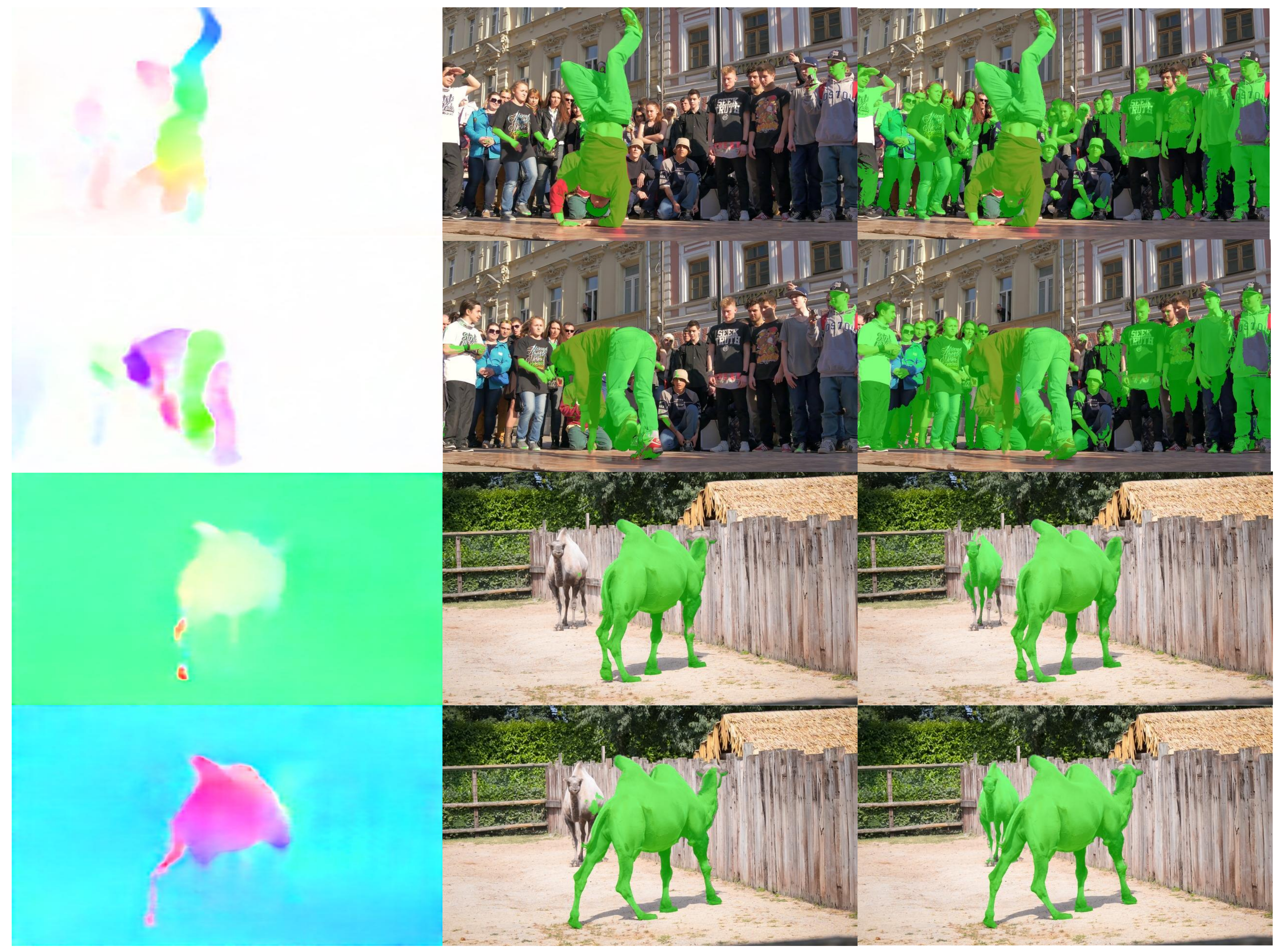}
   \caption{Flow in the left column shows the actual motion of objects. The middle column shows the results after adding supervision to the flow branch which performs much better with fewer false positives (the audience and the camel in the back), compared to the right column which is trained without flow branch supervision.}
\label{fig:daviscompare}
\end{figure}

\subsection{Ablation study}\label{secex-ablation}
We offer to discuss several major considerations when implementing the whole UDA-involved VOS pipeline. The effects of different sub-modules is reflected through mean $\mathcal{J}$ score as presented in Table \ref{davisablation}. We set the model without flow branch as the baseline, which conducts VOS from only static images. The study mainly focus on the fusion strategy, supervision on flow branch and behavior of different UDA methods.  

\textbf{Fusion method}  Table \ref{davisablation} shows that the fusion of appearance and motion cue plays a major role in performance boost. Comparing to the baseline, using flow branch as the motion cue considerably raises the performance. For the fusion layer, we look into several alternatives, which include pixel-wise product, pixel-wise addition, and convolution. The results imply that the learnable module, a.k.a. the convolution layer is better at handling the task, especially when adding supervision on the flow branch which outperforms other operations with $81.8\%$ mean $\mathcal{J}$ score without CRFs. 

\textbf{Flow branch supervision} As we have illustrated in \S \ref{secapp}, the flow supervision monitors the feature extraction by setting up a separated decoder $De^{flow}$ and attach $L_S^{flow}$ to its output $\boldsymbol{Y}_S^{flow}$. Such a strategy aims to make better use of motion cues. It turns out to be effective compared to only pay attention to $\boldsymbol{X}_S$. Table \ref{davisablation} shows that such strategy raises the score of $\mathcal{J}$ by $2.1\%$ after post-process. We also observe that, without supervision on the flow branch, the prediction process tends to only focus on appearance cues. In this case, the model mistakenly captures static objects as presented in Figure \ref{fig:daviscompare}, where the audience in the breakdance scene and the camel in the back are all classified as moving. Such false positive cases significantly decrease after training with supervision on the flow branch.

\textbf{Domain adaptation strategy}
We adopt the idea in \cite{adda} when conducting adversarial UDA in the VOS task. But instead of only using weights-separated setup, we also implement the weights-shared setup, which performs better in this task. Table \ref{fbmsbenckmark} and Table \ref{ytbbenchmark} show that the weights-shared strategy outperforms the weights-separated one on both FBMS59 \cite{fbms} and Youtube-Object \cite{ytb}. The $\mathcal{J}$ score increases by $6.3\%$ on FBMS59 and $10.4\%$ on Youtube-Object after weights-shared UDA. For weights-separated UDA the numbers are $2.3\%$ and $4.0\%$. It shows that domain adaptations becomes more effective when it is conducted along with the training on the source domain, which we assume the alignment of features of different domains behaves more flexible as well as lower the risk of sabotaging the ability of conducting VOS compared to training a separated feature extractor.

\section{Conclusion}\label{conclusion}
We have proposed a method combining supervised learning and unsupervised domain adaptation strategy for video object segmentation on unlabeled data. Through comparing different fusion operations and adding supervision on flow feature extraction during supervised training, we make sure our model distinguish between static and moving foreground. The impact of shared and separated weights in UDA process is also taken care of in our work. Our proposed UDA method manages to maintaining the performance for the model when applied to unlabeled datasets.

{\small
\bibliographystyle{ieee_fullname}
\bibliography{egpaper}
}

\end{document}